\documentclass{aip-cp}

\usepackage[numbers]{natbib}
\usepackage{rotating}
\usepackage{array}
\usepackage{booktabs}
\usepackage{xurl}
\usepackage{multirow}
\usepackage{xcolor}
\usepackage{enumerate}
\usepackage{adjustbox}

\begin{document}

\title{Remote Sensing Imagery for Flood Detection: Exploration of Augmentation Strategies}

\author[aff1,aff2]{Vladyslav Polushko \corref{cor1}}

\author[aff1]{Damjan Hatic}

\author[aff1]{Ronald Rösch}

\author[aff2]{Thomas März}

\author[aff1]{Markus Rauhut}

\author[aff2]{Andreas Weinmann}

\affil[aff1]{Image Processing Department, Fraunhofer ITWM, Kaiserslautern, Germany}
\affil[aff2]{Algorithms for Computer Vision, Imaging and Data Analysis Group, Hochschule Darmstadt, Darmstadt, Germany}
\corresp[cor1]{Vladyslav Polushko: vladyslav.polushko@itwm.fraunhofer.de}

\maketitle

\begin{abstract}
  Floods cause serious problems around the world. Responding quickly and effectively requires accurate and timely information about the affected areas. The effective use of Remote Sensing images for accurate flood detection requires specific detection methods. Typically, Deep Neural Networks are employed, which are trained on specific datasets. For the purpose of river flood detection in RGB imagery, we use the BlessemFlood21 dataset. We here explore the use of different augmentation strategies, ranging from basic approaches to more complex techniques, including optical distortion. By identifying effective strategies, we aim to refine the training process of state-of-the-art Deep Learning segmentation networks.
\end{abstract}

\section{INTRODUCTION}

In 2023, floods were listed among the three disasters with the most fatalities~\cite{ourworld2021online, IPPC21}. These floods frequently cause a significant number of deaths and result in substantial economic losses. Hence, the development of effective response and recovery strategies is of utmost importance.
Imagery from Remote Sensing (RS) provides important data for the depiction and surveillance of flood-impacted areas. Drones are frequently utilized for this purpose due to their ability to deliver high-resolution images, operate in cloudy conditions, and their quick deployment capabilities. These attributes facilitate the accurate mapping of flooded regions for immediate analysis~\cite{gebrehiwot19dlmappingdroneflood}. Moreover, drone imagery contributes to the identification of regions prone to flooding, which aids in post-event flood risk analysis~\cite{UnicefDronesFlood22, karamuz20dronesfloodhazardassessment,munawar21applicationdronesfloodoldandDL}.
Robust Computer Vision (CV) algorithms are needed for managing the substantial image data and enabling prompt decision-making~\cite{gebrehiwot20inundationmappingndwidl, RealTimeFloodDetectionDL21, HumanitarianDronesFloodDeepLearning20, karamuz20dronesfloodhazardassessment}. One key task these algorithms perform is semantic segmentation of water, which consists of classifying each pixel in an image with respect to whether it contains water or not, enabling detailed and precise identification of flooded areas~\cite{minaee21imagesegmentationreview, chandran24deepflooddelination}. A variety of strategies have been formulated for the purpose of water identification. For example, Spectral Indices like the Normalized Difference Water Index~\cite{mcfeeters1996ndwi} are generated using satellite imagery highlighting water.
Deep Learning (DL) techniques have progressively overtaken conventional approaches to flood detection. Deep Convolutional Neural Networks have shown effectiveness in managing large and varied datasets, including satellite imagery. An analysis by Bentivoglio et al.~\cite{bentivoglio21deeplearningforfloodingreview} compared various DL models used in flood mapping, emphasizing their enhanced accuracy and efficiency.
In Hashemi-Beni et al.~\cite{hashemi21floodmappinguav} employed a Fully Convolutional Network to perform semantic segmentation for the detection of flooded areas. The training of these DL models needs the availability of comprehensive data that is task-specific and adequately represents the scenarios for which the models are being trained.
There are datasets in RS that are particularly applicable for semantic segmentation of flood water. For instance, the FloodNet~\cite{rahnemoonfar2021floodnet} dataset includes high-resolution RGB data and nine unique multiclass masks. These originate from the aftermath of Hurricane Harvey and offer a detailed representation at 1.5~cm resolution of the areas that floods impacted. Additionally, the BlessemFlood21 dataset~\cite{polushko24blessem} is specifically curated for semantic water segmentation, featuring RGB imagery at 15 cm spatial resolution of river scenes affected by flooding.

To enhance the performance of DL models, a common strategy is image augmentation~\cite{yang22augmentationsurvey}. Image augmentation is a technique used in image processing to enrich the training set without collecting new data. By applying transformations such as rotations, translations, and more advanced strategies, the diversity of the training set can be increased. This not only helps to prevent overfitting and improves the model's ability to generalize~\cite{yang22augmentationsurvey,minaee21imagesegmentationreview}.

\paragraph{Contributions.}

State-of-the-art methods for water detection, with the exception of~\cite{patel21autoaugmentrivers, alharbi20selectiveaugmentation}, use only a small number of basic augmentations such as rotation or flipping~\cite{ahmad19roadsafterfloods, wang21hawaterbodyaugmentation, muhadi21deepsurveillanceriverflood, peng19urbanfloodmapping, bai21wateraugmentbasic}. We investigate several image augmentations and their influence on model performance, particularly in the context of water segmentation. Our approach involves segregating augmentations into groups and individually applying each group during the training phase to evaluate the effect on model performance. Our research incorporates nine unique augmentation groups and uses the two contemporary models UNet++~\cite{zhou18unet++} and DeepLabV3+~\cite{chen18deeplabv3+}, each undergoing training for 100, 200, and 300 epochs. We employ these augmentations and training procedures using the BlessemFlood21 dataset. Our contributions are as follows: (i) We systematically structured various image augmentations into distinct groups. This structured approach aids in better understanding the influence of each augmentation. (ii) We conduct an evaluation of the influence of each group, measuring the performance of state-of-the-art models like UNet++ and DeepLabV3+, laying a baseline for further research. (iii) In a comparison, we identify advantageous and less effective augmentation groups over a training period of 300 epochs. Moreover, we confirm the need for extended training durations to further enhance the performance of each group.

\section{AUGMENTATIONS}

We examine and compare the influence of various augmentations on the performance of models tailored for the semantic segmentation of flood water. For this purpose, we use the Albumentations library of Buslaev et al.~\cite{buslaev20albumentations}.

\paragraph{Categorization.}
\label{subsec:categorization}

The available augmentations are categorized into nine distinct groups: (i) The \textbf{Basic} augmentation group includes simple transformations such as resizing, flipping, rotating, mirroring, and cropping, all aimed at mimicking a variety of perspectives~\cite{alomar2023data}. (ii) The \textbf{Blur and Noise} augmentation category encompasses techniques that simulate image unsharpness when out of focus, a phenomenon frequently encountered during image acquisition in motion, such as those involving drones or gyrocopters. The group includes several blur variants: Gaussian, Motion, Median, Advanced, and Zoom, employed to simulate conditions of lens defocus. In addition, it offers an element of noise by applying ISO and multiplicative alterations to mirror sensor noise artifacts. (iii) The \textbf{Color}  augmentation group focuses on transformations that manipulate the color and lighting properties of images. This involves techniques like Color Jitter, which introduces randomness in color saturation, brightness, contrast, and hue; Random Gamma, manipulating the gamma levels to achieve varying brightness effects; ToGray, converting colored images to grayscale; Solarize, inverting all pixel values above a threshold; InvertImg, a complete inversion of color mappings; RGBShift, shifting RGB values of the image; and Equalize, balancing the color distribution. These transformations alter the colors as well as their interactions. (iv) The \textbf{Drop} augmentation group integrates more intricate transformations that exclude or modify portions of images. This includes techniques such as Mask Dropout, which randomly sets sections of the input unit to zero; Grid Dropout, inducing structured random holes in the image; Coarse Dropout, randomly setting larger rectangular regions to zero; and CropNonEmptyMaskIfExists, which crops the non-empty part if the mask exists. Also, Random Crop is applied to yield a specified ratio of an image from random borders. This group aims to help augment sensor artifacts as well as occlusions. (v) The \textbf{Distortion}  augmentation category comprises a variety of techniques manipulating an image's spatial structure. Included are methods such as OpticalDistortion that warps the image by applying barrel or pincushion distortion, ElasticTransform which alters an image according to a displacement map using random deformation, GridDistortion to randomly shift pixels in an image to warp the scenery, and GridDropout for removing pixels in a structural manner. In addition, GridElasticDeform applies deformation over a grid to create a complex and elastic transformation. The Perspective technique alters the perspective view of the image, as if taken from a different viewpoint. RandomScale is employed to randomly scale the image within a set range, while CropAndPad crops the image and pads the remaining part. (vi) The \textbf{Pixel}  augmentation group operates around granular, pixel-level modifications, providing a detailed level of image variation. This category incorporates techniques such as RandomToneCurve, which modulates the image's color channels, RingingOverShoot to introduce ringing artifacts, and Sharpen for enhancing edge detail. Specifically, the Emboss transformation emphasizes edges to render a three-dimensional effect. Techniques like ChannelShuffle, which rearranges the image channels randomly, and ChannelDropout, which selectively leaves out certain channels, are also employed. (vii) The \textbf{Quality}  augmentation group focuses on augmentations that manipulate overall image quality. This includes Downscaling to reduce image resolution, ISO Noise to model camera sensor noise, Defocus to blur the image, simulating lens effects, Image Compression to decrease the file size while degrading image quality, and Posterize which reduces the color bit depth, leading to fewer distinct colors in the image and creating a poster-like effect. (viii) The \textbf{Sophisticated}  augmentation group houses advanced augmentation techniques that apply complex adjustments. This includes PlanckianJitter, which shifts the color balance of the image towards warmer or cooler tones, Superpixels, which segments the image into numerous superpixel regions, and FancyPCA for color augmentation that alters the intensities of the RGB channels and finally Contrast Limited Adaptive Histogram Equalization (CLAHE), an adaptive method to enhance image contrast. (ix) Finally, the \textbf{Weather}  augmentation group renders various weather phenomena in images. This includes Random Sun Flare that overlays an artificial sun flare effect, Random Rain and Random Splatter creating the impression of raindrops, Random Fog which imparts a hazy atmosphere, and Random Shadow introducing artificial shadows. These weather imitating augmentations guide models in recognizing targets under different environmental conditions, thus enhancing overall robustness.

\paragraph{Parameters.}

Each augmentation group is normalized to adjust image values within the 2nd and 98th percentile range, maintaining proper contrast. Each augmentation within a group is applied with a certain probability. As a result, in some instances, all augmentations might be applied, while in rare cases, none might be.
The default probability suggested by the Albumentations library of 50\% is applied to all augmentations except for the blur ones, where we chose a probability of 30\% to avoid excessive blurring of images. Each augmentation involves a set of parameters, and we utilized the default Albumentations parameters for all except for a slight modification in the basic group: the functionality of ShiftScaleRotate was divided into separate shift, scale, and rotation functions; we adjusted the default parameters for shifting and scaling from 0.1 to 0.3 and permitted a full 360-degree rotation.

\paragraph{Setup.}

In our study, we choose the BlessemFlood21 dataset~\cite{polushko24blessem} as it is specifically designed for semantic segmentation of floodwater in RGB imagery, offering flooded river scenery high resolution at 15 cm. 
We employ two established semantic segmentation models, DeeplabV3+~\cite{chen18deeplabv3+} and UNet++~\cite{zhou18unet++}, both initially pre-trained on the ImageNet dataset~\cite{deng09imagenet} and subsequently trained on the augmented BlessemFlood21 dataset.
To preserve detail, we use 4623 square tiles, each measuring 512 by 512 pixels. The BlessemFlood21 dataset is divided into 80\% for training, 10\% for validation, and 10\% for testing. In particular, we set aside 10\% of the images containing water for both the validation and test sets to ensure a varied representation across these subsets.
 We conduct the training on an NVIDIA H100 with 80 GB and analyze the performance results using Intersection over Union (IoU), Dice score (Dice) and the Accuracy (Acc). For more details on these metrics, see~\cite{minaee21imagesegmentationreview}.


\section{RESULT AND DISCUSSION}

\paragraph{Performance Improvement.}
\label{subsec:Performance Improvement}

We assess the performance of each augmentation group across 300 training epochs, using the basic group as a benchmark. The summarized performance results can be found in Table~\ref{tab:augmentation_comparison_unet} and Table~\ref{tab:augmentation_comparison_deeplab}. The Table displays remarkable performance across all metrics for each augmentation group, with IoU achieving more than 83\%, the Dice score exceeding 92\%, and Accuracy reaching a minimum of 99.6\%. An important observation is the slight performance variation when juxtaposing each augmentation for both UNet++ and DeepLabV3+ models. The deviation peaks at 2\% for IoU, 1\% for Dice, and less than 0.05\% for Accuracy.
Upon inspecting the IoU and Dice scores, the Distortion augmentation group excels with the UNet++ model, while the Blur and Noise group exhibits slightly enhanced performance with the DeepLabV3+ model. Despite these discrepancies, the differences remain relatively insignificant. The Color augmentation group registers the lowest scores across all metrics for both models; however, the scores remain noteworthy, with IoU above 82\%, Dice above 90\%, and Accuracy exceeding 99.5\%.
Our evaluation indicates that while the choice of augmentation techniques can influence the performance of semantic segmentation models, the extent of variation is relatively minor. Every augmentation group examined displays strong performance across all metrics, with the Color augmentation group, recording the least scores. This observation underlines that employing a variety of augmentation techniques can beneficially impact the performance, particularly for semantic segmentation of floodwater in high-resolution RS imagery such as in BlessemFlood21.

\paragraph{Epoch Influence.}

We investigate the impact of the number of training epochs on the model's performance when using augmentations. Our test consists of three different epoch lengths: 100, 200, and 300, aiming to determine if extended training of 300 epochs is necessary for maximizing performance improvements. The results for each epoch number are presented in Table~\ref{tab:augmentation_comparison_unet} and Table~\ref{tab:augmentation_comparison_deeplab}
respectively.
Our observations indicate that extended training enhances model performance and results in smaller deviations across each augmentation group. 
These findings align with our previous observations, where the Distortion augmentation group most significantly improves the performance of the UNet++ model, and the Blur and Noise augmentation group is most beneficial for the DeepLabV3+ model. 
On the other hand, the Color augmentation group appears to be the least beneficial augmentation for both models across all numbers of epochs. It consistently yields the lowest scores across all metrics, with notably lower values for the DeepLabV3+ model trained for 100 epochs.
In summary, the number of epochs generally enhances the performance of the models across all augmentation groups. The Distortion augmentation and Blur and Noise augmentation groups consistently yield the best performance, while the Color augmentation group tends to lag behind. Nonetheless, even the Color augmentation group, being the least effective, still demonstrates high scores, suggesting that all the augmentation groups contribute positively to some extent.

\begin{table}[t]
  \begin{adjustbox}{width=0.85\textwidth,center}
  \centering
  \begin{tabular}{p{2.1cm}|p{1.15cm}|p{1.26cm}|p{1.17cm}|p{1.2cm}|p{1.26cm}|p{1.2cm}|p{1.15cm}|p{1.26cm}|p{1.2cm}}
  \toprule
  Model & \multicolumn{9}{c|}{UNet++}   \\
  \midrule
  Epoch & \multicolumn{3}{c|}{100} & \multicolumn{3}{c|}{200} & \multicolumn{3}{c}{300} \\
  \midrule
   & IoU (\%) & Dice (\%) & Acc (\%) & IoU (\%) & Dice (\%) & Acc (\%) & IoU (\%) & Dice (\%) & Acc (\%) \\
  \midrule
  Basic & 88.69 & 94.00 & 99.72 & 89.11 & 94.24 & 99.73 & 88.89 & 94.12 & 99.73 \\ 
  Blur and Noise & 88.29 & 93.78 & 99.71 & \underline{89.87} & \underline{94.66} & \underline{99.75} & \underline{89.78} & \underline{94.61} & \underline{99.75} \\ 
  Color & 82.55 & 90.44 & 99.56 & 84.22 & 91.10 & 99.60 & 83.61 & 91.08 & 99.60\\ 
  Drop & 88.21 & 93.73 & 99.71 & 89.12 & 94.25 & 99.73 & 89.12 & 94.25 & 99.73\\ 
  Distortion & \textbf{90.21} & \textbf{94.85} & \textbf{99.76} & \textbf{90.22} & \textbf{94.86} & \textbf{99.76} & \textbf{90.56} & \textbf{95.05} & \textbf{99.79}\\ 
  Pixel & 87.69 & 93.44 & 99.70 & 87.23 & 93.18 & 99.69 & 85.86 & 92.39 & 99.65 \\ 
  Quality & 86.89 & 92.96 & 99.68 & 89.32 & 94.36 & 99.74 & 89.00 & 94.19 & 99.73\\ 
  Sophisticated & \underline{88.90} & \underline{94.13} & \underline{99.72} & 88.92 & 94.14 & 99.73 & 88.72 & 94.02 & 99.72\\ 
  Weather & 84.95 & 91.86 & 99.61 & 86.71 & 92.88 & 99.66 & 86.71 & 92.88 & 99.67\\  
  \bottomrule
  \end{tabular}
  \caption{Performance comparison of nine augmentation groups trained on BlessemFlood21 using UNet++ for 100, 200, and 300 epochs. In each epoch, bold values indicate top performance and underlined values denote the second best.}
  \label{tab:augmentation_comparison_unet}
\end{adjustbox}
  \end{table}

  \begin{table}[t]
    \begin{adjustbox}{width=0.85\textwidth,center}
    \centering
    \begin{tabular}{p{2.1cm}|p{1.15cm}|p{1.26cm}|p{1.17cm}|p{1.2cm}|p{1.26cm}|p{1.2cm}|p{1.15cm}|p{1.26cm}|p{1.2cm}}
    \toprule
    Model & \multicolumn{9}{c|}{DeepLabV3+}   \\
    \midrule
    Epoch & \multicolumn{3}{c|}{100} & \multicolumn{3}{c|}{200} & \multicolumn{3}{c}{300} \\
    \midrule
     & IoU (\%) & Dice (\%) & Acc (\%) & IoU (\%) & Dice (\%) & Acc (\%) & IoU (\%) & Dice (\%) & Acc (\%) \\
    \midrule
    Basic & 80.93 & 89.46 & 99.54 & 88.53 & 93.92 & 99.72 & 88.69 & 94.00 & 99.72\\ 
    Blur and Noise & \textbf{88.44} & \textbf{93.87} & \textbf{99.72} & \textbf{88.68} & \textbf{94.00} & \textbf{99.72} & \underline{88.75} & \underline{94.04} & \underline{99.73} \\ 
    Color & 10.68 & 19.3 & 97.59 & 83.45 & 90.98 & 99.59 & 82.88 & 90.64 & 99.58\\ 
    Drop & 87.09 & 93.10 & 99.67 & 86.72 & 92.89 & 99.67 & 88.44 & 93.89 & 99.71 \\ 
    Distortion & \underline{87.52} & \underline{93.95} & \underline{99.72} & \underline{88.59} & \underline{93.95} & \underline{99.72} & \textbf{88.93} & \textbf{94.14} & \textbf{99.73}\\ 
    Pixel & 72.83 & 84.32 & 99.35 & 86.14 & 92.56 & 99.66 & 86.88 & 92.98 & 99.68 \\ 
    Quality & 76.71 & 86.21 & 99.43 & 87.66 & 93.43 & 99.69 & 87.66 & 93.43 & 99.69\\ 
    Sophisticated & 81.68 & 89.92 & 99.54  & 88.24 & 93.75 & 99.71 & 88.24 & 93.75 & 99.71\\ 
    Weather & 85.12 & 91.96 & 99.63 & 83.56 & 91.04 & 99.58 & 86.44 & 92.73 & 99.67\\  
    \bottomrule
    \end{tabular}
    \caption{Performance comparison of nine augmentation groups trained on BlessemFlood21 using DeepLabV3+ for 100, 200, and 300 epochs. In each epoch, bold values indicate top performance and underlined values denote the second best.}
    \label{tab:augmentation_comparison_deeplab}
  \end{adjustbox}
  \end{table}

\section{CONCLUSION}

We investigated the impact of various image augmentation techniques on the performance of DL models for semantic segmentation of flood imagery. We assigned available augmentations into nine distinct groups and evaluated their effectiveness using the state-of-the-art segmentation models UNet++ and DeepLabV3+ over different training lengths (100, 200, and 300 epochs). We found that certain augmentations, in particular the Distortion augmentation group and the Blur and Noise augmentation group, enhance model performance the most, and that longer training periods generally improve model performance. Our findings provide a structured framework for incorporating image augmentation into model training for flood detection. Future research could explore combining augmentation groups and optimizing hyperparameters, such as transformation application probabilities, to potentially enhance model performance.

\section{ACKNOWLEDGMENTS}
The authors gratefully acknowledge the computing time provided on the high-performance computer Lichtenberg II at TU Darmstadt, funded by the German Federal Ministry of Education and Research (BMBF) and the State of Hesse.

\bibliographystyle{aipnum-cp}
\bibliography{sample}%

\begin{thebibliography}{29}%
\makeatletter
\providecommand \@ifxundefined [1]{%
 \@ifx{#1\undefined}
}%
\providecommand \@ifnum [1]{%
 \ifnum #1\expandafter \@firstoftwo
 \else \expandafter \@secondoftwo
 \fi
}%
\providecommand \@ifx [1]{%
 \ifx #1\expandafter \@firstoftwo
 \else \expandafter \@secondoftwo
 \fi
}%
\providecommand \natexlab [1]{#1}%
\providecommand \enquote  [1]{``#1''}%
\providecommand \bibnamefont  [1]{#1}%
\providecommand \bibfnamefont [1]{#1}%
\providecommand \citenamefont [1]{#1}%
\providecommand \href@noop [0]{\@secondoftwo}%
\providecommand \href [0]{\begingroup \@sanitize@url \@href}%
\providecommand \@href[1]{\@@startlink{#1}\@@href}%
\providecommand \@@href[1]{\endgroup#1\@@endlink}%
\providecommand \@sanitize@url [0]{\catcode `\$12\catcode `\&12\catcode
  `\#12\catcode `\^12\catcode `\_12\catcode `\%12\relax}%
\providecommand \@@startlink[1]{}%
\providecommand \@@endlink[0]{}%
\providecommand \url  [0]{\begingroup\@sanitize@url \@url }%
\providecommand \@url [1]{\endgroup\@href {#1}{\urlprefix }}%
\providecommand \urlprefix  [0]{URL }%
\providecommand \Eprint [0]{\href }%
\providecommand \doibase [0]{http://dx.doi.org/}%
\providecommand \selectlanguage [0]{\@gobble}%
\providecommand \bibinfo  [0]{\@secondoftwo}%
\providecommand \bibfield  [0]{\@secondoftwo}%
\providecommand \translation [1]{[#1]}%
\providecommand \BibitemOpen [0]{}%
\providecommand \bibitemStop [0]{}%
\providecommand \bibitemNoStop [0]{.\EOS\space}%
\providecommand \EOS [0]{\spacefactor3000\relax}%
\providecommand \BibitemShut  [1]{\csname bibitem#1\endcsname}%
\let\auto@bib@innerbib\@empty
\bibitem [{\citenamefont {Ritchie}\ and\ \citenamefont
  {Rosado}(2022)}]{ourworld2021online}%
  \BibitemOpen
  \bibfield  {author} {\bibinfo {author} {\bibfnamefont {H.}~\bibnamefont
  {Ritchie}}\ and\ \bibinfo {author} {\bibfnamefont {P.}~\bibnamefont
  {Rosado}},\ }\href@noop {} {\bibinfo {title} {Natural disasters},\ }\
  \bibinfo {howpublished} {Our World in Data, 2022,
  https://ourworldindata.org/natural-disasters (accessed June 22, 2024)} (
  \bibinfo {year} {2022} \unskip)\BibitemShut {NoStop}%
\bibitem [{\citenamefont {Arias}\ \emph {et~al.}(2021)\citenamefont {Arias},
  \citenamefont {Bellouin}, \citenamefont {Coppola}, \citenamefont {Jones},
  \citenamefont {Krinner}, \citenamefont {Gerhard},\ and\ \citenamefont
  {Marotzke~et al.}}]{IPPC21}%
  \BibitemOpen
  \bibfield  {author} {\bibinfo {author} {\bibfnamefont {P.}~\bibnamefont
  {Arias}}, \bibinfo {author} {\bibfnamefont {N.}~\bibnamefont {Bellouin}},
  \bibinfo {author} {\bibfnamefont {E.}~\bibnamefont {Coppola}}, \bibinfo
  {author} {\bibfnamefont {R.}~\bibnamefont {Jones}}, \bibinfo {author}
  {\bibnamefont {Krinner}}, \bibinfo {author} {\bibnamefont {Gerhard}}, \ and\
  \bibinfo {author} {\bibfnamefont {J.}~\bibnamefont {Marotzke~et al.}},\
  }\href {https://www.ipcc.ch/report/ar6/wg1/} {\enquote {\bibinfo {title}
  {{2021: Technical Summary. In Climate Change 2021: The Physical Science
  Basis. Contribution of Working Group I to the Sixth Assessment Report of the
  Intergovernmental Panel on Climate Change}},}\ }\bibinfo {type} {Tech. Rep.}\
  (\bibinfo  {institution} {IPPC},\ \bibinfo {address} {United Kingdom and New
  York},\ \bibinfo {year} {2021})\ \bibinfo {note}
  {\url{https://www.ipcc.ch/report/ar6/wg1/downloads/report/IPCC_AR6_WGI_TS.pdf}}\BibitemShut
  {NoStop}%
\bibitem [{\citenamefont {Gebrehiwot}\ \emph {et~al.}(2019)\citenamefont
  {Gebrehiwot}, \citenamefont {Hashemi-Beni}, \citenamefont {Thompson},
  \citenamefont {Kordjamshidi},\ and\ \citenamefont
  {Langan}}]{gebrehiwot19dlmappingdroneflood}%
  \BibitemOpen
\bibinfo {author} {\bibfnamefont {A.}~\bibnamefont {Gebrehiwot}}, \bibinfo
  {author} {\bibfnamefont {L.}~\bibnamefont {Hashemi-Beni}}, \bibinfo {author}
  {\bibfnamefont {G.}~\bibnamefont {Thompson}}, \bibinfo {author}
  {\bibfnamefont {P.}~\bibnamefont {Kordjamshidi}}, \ and\ \bibinfo {author}
  {\bibfnamefont {T.~E.}\ \bibnamefont {Langan}}\bibfield  {author} {
  }\enquote {\bibinfo {title} {Deep convolutional neural network for flood
  extent mapping using unmanned aerial vehicles data},}\ ,\ \href@noop {}
  {\bibfield  {journal} {\bibinfo  {journal} {Sensors}\ }\textbf {\bibinfo
  {volume} {19}},\ p.\ \bibinfo {pages} {1486} (\bibinfo {year}
  {2019})}\BibitemShut {NoStop}%
\bibitem [{\citenamefont {{UNICEF Malawi}}(2021)}]{UnicefDronesFlood22}%
  \BibitemOpen
  \bibfield  {author} {\bibinfo {author} {\bibnamefont {{UNICEF Malawi}}},\
  }\href@noop {} {\bibinfo {title} {{Summative Evaluation of the Impact of
  Using Drones on Population Health \& Other Outcomes}},\ }\ \bibinfo
  {howpublished}
  {\url{https://www.unicef.org/malawi/reports/summative-evaluation-impact-using-drones-population-health-other-outcomes}}
  ( \bibinfo {year} {2021} \unskip),\ \bibinfo {note} {accessed:
  2023-07-07}\BibitemShut {NoStop}%
\bibitem [{\citenamefont {Karamuz}, \citenamefont {Romanowicz},\ and\
  \citenamefont {Doroszkiewicz}(2020)}]{karamuz20dronesfloodhazardassessment}%
  \BibitemOpen
\bibinfo {author} {\bibfnamefont {E.}~\bibnamefont {Karamuz}}, \bibinfo
  {author} {\bibfnamefont {R.~J.}\ \bibnamefont {Romanowicz}}, \ and\ \bibinfo
  {author} {\bibfnamefont {J.}~\bibnamefont {Doroszkiewicz}}\bibfield  {author}
  {  }\enquote {\bibinfo {title} {The use of unmanned aerial vehicles in flood
  hazard assessment},}\ ,\ \href@noop {} {\bibfield  {journal} {\bibinfo
  {journal} {Journal of Flood Risk Management}\ }\textbf {\bibinfo {volume}
  {13}},\ p.\ \bibinfo {pages} {e12622} (\bibinfo {year} {2020})}\BibitemShut
  {NoStop}%
\bibitem [{\citenamefont {Munawar}\ \emph {et~al.}(2021)\citenamefont
  {Munawar}, \citenamefont {Ullah}, \citenamefont {Qayyum},\ and\ \citenamefont
  {Heravi}}]{munawar21applicationdronesfloodoldandDL}%
  \BibitemOpen
\bibinfo {author} {\bibfnamefont {H.~S.}\ \bibnamefont {Munawar}}, \bibinfo
  {author} {\bibfnamefont {F.}~\bibnamefont {Ullah}}, \bibinfo {author}
  {\bibfnamefont {S.}~\bibnamefont {Qayyum}}, \ and\ \bibinfo {author}
  {\bibfnamefont {A.}~\bibnamefont {Heravi}}\bibfield  {author} {  }\enquote
  {\bibinfo {title} {Application of deep learning on uav-based aerial images
  for flood detection},}\ ,\ \href@noop {} {\bibfield  {journal} {\bibinfo
  {journal} {Smart Cities}\ }\textbf {\bibinfo {volume} {4}},\ \unskip\
  \bibinfo {pages} {1220--1242} (\bibinfo {year} {2021})}\BibitemShut {NoStop}%
\bibitem [{\citenamefont {Gebrehiwot}\ and\ \citenamefont
  {Hashemi-Beni}(2020)}]{gebrehiwot20inundationmappingndwidl}%
  \BibitemOpen
  \bibfield  {author} {\bibinfo {author} {\bibfnamefont {A.}~\bibnamefont
  {Gebrehiwot}}\ and\ \bibinfo {author} {\bibfnamefont {L.}~\bibnamefont
  {Hashemi-Beni}},\ }\enquote {\bibinfo {title} {{Automated indunation mapping:
  comparison of methods}},}\ in\ \href@noop {} {\emph {\bibinfo {booktitle}
  {IGARSS 2020 IEEE International Geoscience and Remote Sensing Symposium}}}\
  (\bibinfo {year} {2020})\ \unskip, pp.\ \bibinfo {pages}
  {3265--3268}\BibitemShut {NoStop}%
\bibitem [{\citenamefont {Hashi}\ \emph {et~al.}(2021)\citenamefont {Hashi},
  \citenamefont {Abdirahman}, \citenamefont {Elmi}, \citenamefont {Hashi},\
  and\ \citenamefont {Rodriguez}}]{RealTimeFloodDetectionDL21}%
  \BibitemOpen
\bibinfo {author} {\bibfnamefont {A.}~\bibnamefont {Hashi}}, \bibinfo {author}
  {\bibfnamefont {A.}~\bibnamefont {Abdirahman}}, \bibinfo {author}
  {\bibfnamefont {M.}~\bibnamefont {Elmi}}, \bibinfo {author} {\bibfnamefont
  {S.}~\bibnamefont {Hashi}}, \ and\ \bibinfo {author} {\bibfnamefont
  {O.}~\bibnamefont {Rodriguez}}\bibfield  {author} {  }\enquote {\bibinfo
  {title} {{A Real-Time Flood Detection System Based on Machine Learning
  Algorithms with Emphasis on Deep Learning}},}\ ,\ \href {\doibase
  10.14445/22315381/IJETT-V69I5P232} {\bibfield  {journal} {\bibinfo  {journal}
  {International Journal of Engineering Trends and Technology}\ }\textbf
  {\bibinfo {volume} {69}},\ \unskip\ \bibinfo {pages} {249--256}05 (\bibinfo
  {year} {2021})}\BibitemShut {NoStop}%
\bibitem [{\citenamefont {Oren}\ and\ \citenamefont
  {Verity}(2020)}]{HumanitarianDronesFloodDeepLearning20}%
  \BibitemOpen
  \bibfield  {author} {\bibinfo {author} {\bibfnamefont {C.}~\bibnamefont
  {Oren}}\ and\ \bibinfo {author} {\bibfnamefont {A.}~\bibnamefont {Verity}},\
  }\href
  {https://www.digitalhumanitarians.com/artifical_intelligence_applied_to_uavs/}
  {\enquote {\bibinfo {title} {{Artifical Intelligence (AI) Applied to Unmanned
  Aerial Vehicles (UAVs) And its Impact on Humanitarian Action}},}\ }\bibinfo
  {type} {Tech. Rep.}\ (\bibinfo  {institution} {UN-OCHA},\ \bibinfo {year}
  {2020})\BibitemShut {NoStop}%
\bibitem [{\citenamefont {Minaee}\ \emph {et~al.}(2021)\citenamefont {Minaee},
  \citenamefont {Boykov}, \citenamefont {Porikli}, \citenamefont {Plaza},
  \citenamefont {Kehtarnavaz},\ and\ \citenamefont
  {Terzopoulos}}]{minaee21imagesegmentationreview}%
  \BibitemOpen
\bibinfo {author} {\bibfnamefont {S.}~\bibnamefont {Minaee}}, \bibinfo {author}
  {\bibfnamefont {Y.}~\bibnamefont {Boykov}}, \bibinfo {author} {\bibfnamefont
  {F.}~\bibnamefont {Porikli}}, \bibinfo {author} {\bibfnamefont
  {A.}~\bibnamefont {Plaza}}, \bibinfo {author} {\bibfnamefont
  {N.}~\bibnamefont {Kehtarnavaz}}, \ and\ \bibinfo {author} {\bibfnamefont
  {D.}~\bibnamefont {Terzopoulos}}\bibfield  {author} {  }\enquote {\bibinfo
  {title} {Image segmentation using deep learning: A survey},}\ ,\ \href@noop
  {} {\bibfield  {journal} {\bibinfo  {journal} {IEEE transactions on pattern
  analysis and machine intelligence}\ }\textbf {\bibinfo {volume} {44}},\
  \unskip\ \bibinfo {pages} {3523--3542} (\bibinfo {year} {2021})}\BibitemShut
  {NoStop}%
\bibitem [{\citenamefont {Chandran}\ \emph {et~al.}(2024)\citenamefont
  {Chandran}, \citenamefont {Anitha}, \citenamefont {Anusree},\ and\
  \citenamefont {Nair}}]{chandran24deepflooddelination}%
  \BibitemOpen
  \bibfield  {author} {\bibinfo {author} {\bibfnamefont {D.~V.}\ \bibnamefont
  {Chandran}}, \bibinfo {author} {\bibfnamefont {J.}~\bibnamefont {Anitha}},
  \bibinfo {author} {\bibfnamefont {A.}~\bibnamefont {Anusree}}, \ and\
  \bibinfo {author} {\bibfnamefont {G.}~\bibnamefont {Nair}},\ }\enquote
  {\bibinfo {title} {Deep learning-based flood detection system using semantic
  segmentation},}\ in\ \href@noop {} {\emph {\bibinfo {booktitle} {2024 7th
  International Conference on Circuit Power and Computing Technologies
  (ICCPCT)}}},\ Vol.~\bibinfo {volume} {1}\ (\bibinfo {organization} {IEEE},\
  \bibinfo {year} {2024})\ \unskip, pp.\ \bibinfo {pages}
  {1584--1592}\BibitemShut {NoStop}%
\bibitem [{\citenamefont {McFeeters}(1996)}]{mcfeeters1996ndwi}%
  \BibitemOpen
\bibinfo {author} {\bibfnamefont {S.~K.}\ \bibnamefont {McFeeters}}\bibfield
  {author} {  }\enquote {\bibinfo {title} {The use of the normalized difference
  water index (ndwi) in the delineation of open water features},}\ ,\
  \href@noop {} {\bibfield  {journal} {\bibinfo  {journal} {International
  journal of remote sensing}\ }\textbf {\bibinfo {volume} {17}},\ \unskip\
  \bibinfo {pages} {1425--1432} (\bibinfo {year} {1996})}\BibitemShut {NoStop}%
\bibitem [{\citenamefont {Bentivoglio}\ \emph {et~al.}(2021)\citenamefont
  {Bentivoglio}, \citenamefont {Isufi}, \citenamefont {Jonkman},\ and\
  \citenamefont {Taormina}}]{bentivoglio21deeplearningforfloodingreview}%
  \BibitemOpen
\bibinfo {author} {\bibfnamefont {R.}~\bibnamefont {Bentivoglio}}, \bibinfo
  {author} {\bibfnamefont {E.}~\bibnamefont {Isufi}}, \bibinfo {author}
  {\bibfnamefont {S.~N.}\ \bibnamefont {Jonkman}}, \ and\ \bibinfo {author}
  {\bibfnamefont {R.}~\bibnamefont {Taormina}}\bibfield  {author} {  }\enquote
  {\bibinfo {title} {Deep learning methods for flood mapping: A review of
  existing applications and future research directions},}\ ,\ \href@noop {}
  {\bibfield  {journal} {\bibinfo  {journal} {Hydrology and Earth System
  Sciences Discussions}\ }\textbf {\bibinfo {volume} {2021}},\ \unskip\
  \bibinfo {pages} {1--43} (\bibinfo {year} {2021})}\BibitemShut {NoStop}%
\bibitem [{\citenamefont {Hashemi-Beni}\ and\ \citenamefont
  {Gebrehiwot}(2021)}]{hashemi21floodmappinguav}%
  \BibitemOpen
\bibinfo {author} {\bibfnamefont {L.}~\bibnamefont {Hashemi-Beni}}\ and\
  \bibinfo {author} {\bibfnamefont {A.~A.}\ \bibnamefont {Gebrehiwot}}\bibfield
   {author} {  }\enquote {\bibinfo {title} {{Flood extent mapping: an
  integrated method using deep learning and region growing using UAV optical
  data}},}\ ,\ \href@noop {} {\bibfield  {journal} {\bibinfo  {journal} {IEEE
  Journal of Selected Topics in Applied Earth Observations and Remote Sensing}\
  }\textbf {\bibinfo {volume} {14}},\ \unskip\ \bibinfo {pages} {2127--2135}
  (\bibinfo {year} {2021})}\BibitemShut {NoStop}%
\bibitem [{\citenamefont {Rahnemoonfar}\ \emph {et~al.}(2021)\citenamefont
  {Rahnemoonfar}, \citenamefont {Chowdhury}, \citenamefont {Sarkar},
  \citenamefont {Varshney}, \citenamefont {Yari},\ and\ \citenamefont
  {Murphy}}]{rahnemoonfar2021floodnet}%
  \BibitemOpen
\bibinfo {author} {\bibfnamefont {M.}~\bibnamefont {Rahnemoonfar}}, \bibinfo
  {author} {\bibfnamefont {T.}~\bibnamefont {Chowdhury}}, \bibinfo {author}
  {\bibfnamefont {A.}~\bibnamefont {Sarkar}}, \bibinfo {author} {\bibfnamefont
  {D.}~\bibnamefont {Varshney}}, \bibinfo {author} {\bibfnamefont
  {M.}~\bibnamefont {Yari}}, \ and\ \bibinfo {author} {\bibfnamefont {R.~R.}\
  \bibnamefont {Murphy}}\bibfield  {author} {  }\enquote {\bibinfo {title}
  {Floodnet: A high resolution aerial imagery dataset for post flood scene
  understanding},}\ ,\ \href@noop {} {\bibfield  {journal} {\bibinfo  {journal}
  {IEEE Access}\ }\textbf {\bibinfo {volume} {9}},\ \unskip\ \bibinfo {pages}
  {89644--89654} (\bibinfo {year} {2021})}\BibitemShut {NoStop}%
\bibitem [{\citenamefont {Polushko}\ \emph {et~al.}(2024)\citenamefont
  {Polushko}, \citenamefont {Jenal}, \citenamefont {Bongartz}, \citenamefont
  {Weber}, \citenamefont {Hatic}, \citenamefont {R{\"o}sch}, \citenamefont
  {M{\"a}rz}, \citenamefont {Rauhut},\ and\ \citenamefont
  {Weinmann}}]{polushko24blessem}%
  \BibitemOpen
  \bibfield  {author} {\bibinfo {author} {\bibfnamefont {V.}~\bibnamefont
  {Polushko}}, \bibinfo {author} {\bibfnamefont {A.}~\bibnamefont {Jenal}},
  \bibinfo {author} {\bibfnamefont {J.}~\bibnamefont {Bongartz}}, \bibinfo
  {author} {\bibfnamefont {I.}~\bibnamefont {Weber}}, \bibinfo {author}
  {\bibfnamefont {D.}~\bibnamefont {Hatic}}, \bibinfo {author} {\bibfnamefont
  {R.}~\bibnamefont {R{\"o}sch}}, \bibinfo {author} {\bibfnamefont
  {T.}~\bibnamefont {M{\"a}rz}}, \bibinfo {author} {\bibfnamefont
  {M.}~\bibnamefont {Rauhut}}, \ and\ \bibinfo {author} {\bibfnamefont
  {A.}~\bibnamefont {Weinmann}},\ }\enquote {\bibinfo {title} {Blessemflood21:
  Advancing flood analysis with a high-resolution georeferenced dataset for
  humanitarian aid support},}\ in\ \href@noop {} {\emph {\bibinfo {booktitle}
  {IEEE International Geoscience and Remote Sensing Symposium (IGARSS)}}}\
  (\bibinfo {year} {2024})\BibitemShut {NoStop}%
\bibitem [{\citenamefont {Yang}\ \emph {et~al.}(2022)\citenamefont {Yang},
  \citenamefont {Xiao}, \citenamefont {Zhang}, \citenamefont {Guo},
  \citenamefont {Zhao},\ and\ \citenamefont {Shen}}]{yang22augmentationsurvey}%
  \BibitemOpen
\bibinfo {author} {\bibfnamefont {S.}~\bibnamefont {Yang}}, \bibinfo {author}
  {\bibfnamefont {W.}~\bibnamefont {Xiao}}, \bibinfo {author} {\bibfnamefont
  {M.}~\bibnamefont {Zhang}}, \bibinfo {author} {\bibfnamefont
  {S.}~\bibnamefont {Guo}}, \bibinfo {author} {\bibfnamefont {J.}~\bibnamefont
  {Zhao}}, \ and\ \bibinfo {author} {\bibfnamefont {F.}~\bibnamefont
  {Shen}}\bibfield  {author} {  }\enquote {\bibinfo {title} {Image data
  augmentation for deep learning: A survey},}\ ,\ \href@noop {} {\bibfield
  {journal} {\bibinfo  {journal} {arXiv preprint arXiv:2204.08610}\ } (\bibinfo
  {year} {2022})}\BibitemShut {NoStop}%
\bibitem [{\citenamefont {Patel}, \citenamefont {Sharma},\ and\ \citenamefont
  {Gulshan}(2021)}]{patel21autoaugmentrivers}%
  \BibitemOpen
\bibinfo {author} {\bibfnamefont {C.}~\bibnamefont {Patel}}, \bibinfo {author}
  {\bibfnamefont {S.}~\bibnamefont {Sharma}}, \ and\ \bibinfo {author}
  {\bibfnamefont {V.}~\bibnamefont {Gulshan}}\bibfield  {author} {  }\enquote
  {\bibinfo {title} {Evaluating self and semi-supervised methods for remote
  sensing segmentation tasks},}\ ,\ \href@noop {} {\bibfield  {journal}
  {\bibinfo  {journal} {arXiv preprint arXiv:2111.10079}\ } (\bibinfo {year}
  {2021})}\BibitemShut {NoStop}%
\bibitem [{\citenamefont {Alharbi}, \citenamefont {Alhichri},\ and\
  \citenamefont {Bazi}(2020)}]{alharbi20selectiveaugmentation}%
  \BibitemOpen
  \bibfield  {author} {\bibinfo {author} {\bibfnamefont {R.}~\bibnamefont
  {Alharbi}}, \bibinfo {author} {\bibfnamefont {H.}~\bibnamefont {Alhichri}}, \
  and\ \bibinfo {author} {\bibfnamefont {Y.}~\bibnamefont {Bazi}},\ }\enquote
  {\bibinfo {title} {Selective data augmentation approach for remote sensing
  scene classification},}\ in\ \href@noop {} {\emph {\bibinfo {booktitle} {2020
  2nd International Conference on Computer and Information Sciences (ICCIS)}}}\
  (\bibinfo {organization} {IEEE},\ \bibinfo {year} {2020})\ \unskip, pp.\
  \bibinfo {pages} {1--4}\BibitemShut {NoStop}%
\bibitem [{\citenamefont {Ahmad}\ \emph {et~al.}(2019)\citenamefont {Ahmad},
  \citenamefont {Pogorelov}, \citenamefont {Riegler}, \citenamefont
  {Ostroukhova}, \citenamefont {Halvorsen}, \citenamefont {Conci},\ and\
  \citenamefont {Dahyot}}]{ahmad19roadsafterfloods}%
  \BibitemOpen
\bibinfo {author} {\bibfnamefont {K.}~\bibnamefont {Ahmad}}, \bibinfo {author}
  {\bibfnamefont {K.}~\bibnamefont {Pogorelov}}, \bibinfo {author}
  {\bibfnamefont {M.}~\bibnamefont {Riegler}}, \bibinfo {author} {\bibfnamefont
  {O.}~\bibnamefont {Ostroukhova}}, \bibinfo {author} {\bibfnamefont
  {P.}~\bibnamefont {Halvorsen}}, \bibinfo {author} {\bibfnamefont
  {N.}~\bibnamefont {Conci}}, \ and\ \bibinfo {author} {\bibfnamefont
  {R.}~\bibnamefont {Dahyot}}\bibfield  {author} {  }\enquote {\bibinfo {title}
  {Automatic detection of passable roads after floods in remote sensed and
  social media data},}\ ,\ \href@noop {} {\bibfield  {journal} {\bibinfo
  {journal} {Signal Processing: Image Communication}\ }\textbf {\bibinfo
  {volume} {74}},\ \unskip\ \bibinfo {pages} {110--118} (\bibinfo {year}
  {2019})}\BibitemShut {NoStop}%
\bibitem [{\citenamefont {Wang}, \citenamefont {Gao},\ and\ \citenamefont
  {Zhang}(2021)}]{wang21hawaterbodyaugmentation}%
  \BibitemOpen
\bibinfo {author} {\bibfnamefont {Z.}~\bibnamefont {Wang}}, \bibinfo {author}
  {\bibfnamefont {X.}~\bibnamefont {Gao}}, \ and\ \bibinfo {author}
  {\bibfnamefont {Y.}~\bibnamefont {Zhang}}\bibfield  {author} {  }\enquote
  {\bibinfo {title} {Ha-net: A lake water body extraction network based on
  hybrid-scale attention and transfer learning},}\ ,\ \href@noop {} {\bibfield
  {journal} {\bibinfo  {journal} {Remote Sensing}\ }\textbf {\bibinfo {volume}
  {13}},\ p.\ \bibinfo {pages} {4121} (\bibinfo {year} {2021})}\BibitemShut
  {NoStop}%
\bibitem [{\citenamefont {Muhadi}\ \emph {et~al.}(2021)\citenamefont {Muhadi},
  \citenamefont {Abdullah}, \citenamefont {Bejo}, \citenamefont {Mahadi},\ and\
  \citenamefont {Mijic}}]{muhadi21deepsurveillanceriverflood}%
  \BibitemOpen
\bibinfo {author} {\bibfnamefont {N.~A.}\ \bibnamefont {Muhadi}}, \bibinfo
  {author} {\bibfnamefont {A.~F.}\ \bibnamefont {Abdullah}}, \bibinfo {author}
  {\bibfnamefont {S.~K.}\ \bibnamefont {Bejo}}, \bibinfo {author}
  {\bibfnamefont {M.~R.}\ \bibnamefont {Mahadi}}, \ and\ \bibinfo {author}
  {\bibfnamefont {A.}~\bibnamefont {Mijic}}\bibfield  {author} {  }\enquote
  {\bibinfo {title} {Deep learning semantic segmentation for water level
  estimation using surveillance camera},}\ ,\ \href@noop {} {\bibfield
  {journal} {\bibinfo  {journal} {Applied Sciences}\ }\textbf {\bibinfo
  {volume} {11}},\ p.\ \bibinfo {pages} {9691} (\bibinfo {year}
  {2021})}\BibitemShut {NoStop}%
\bibitem [{\citenamefont {Peng}\ \emph {et~al.}(2019)\citenamefont {Peng},
  \citenamefont {Liu}, \citenamefont {Meng},\ and\ \citenamefont
  {Huang}}]{peng19urbanfloodmapping}%
  \BibitemOpen
  \bibfield  {author} {\bibinfo {author} {\bibfnamefont {B.}~\bibnamefont
  {Peng}}, \bibinfo {author} {\bibfnamefont {X.}~\bibnamefont {Liu}}, \bibinfo
  {author} {\bibfnamefont {Z.}~\bibnamefont {Meng}}, \ and\ \bibinfo {author}
  {\bibfnamefont {Q.}~\bibnamefont {Huang}},\ }\enquote {\bibinfo {title}
  {Urban flood mapping with residual patch similarity learning},}\ in\
  \href@noop {} {\emph {\bibinfo {booktitle} {Proceedings of the 3rd ACM
  SIGSPATIAL International Workshop on AI for Geographic Knowledge
  Discovery}}}\ (\bibinfo {year} {2019})\ \unskip, pp.\ \bibinfo {pages}
  {40--47}\BibitemShut {NoStop}%
\bibitem [{\citenamefont {Bai}\ \emph {et~al.}(2021)\citenamefont {Bai},
  \citenamefont {Wu}, \citenamefont {Yang}, \citenamefont {Yu}, \citenamefont
  {Zhao}, \citenamefont {Liu}, \citenamefont {Yang}, \citenamefont {Mas},\ and\
  \citenamefont {Koshimura}}]{bai21wateraugmentbasic}%
  \BibitemOpen
\bibinfo {author} {\bibfnamefont {Y.}~\bibnamefont {Bai}}, \bibinfo {author}
  {\bibfnamefont {W.}~\bibnamefont {Wu}}, \bibinfo {author} {\bibfnamefont
  {Z.}~\bibnamefont {Yang}}, \bibinfo {author} {\bibfnamefont {J.}~\bibnamefont
  {Yu}}, \bibinfo {author} {\bibfnamefont {B.}~\bibnamefont {Zhao}}, \bibinfo
  {author} {\bibfnamefont {X.}~\bibnamefont {Liu}}, \bibinfo {author}
  {\bibfnamefont {H.}~\bibnamefont {Yang}}, \bibinfo {author} {\bibfnamefont
  {E.}~\bibnamefont {Mas}}, \ and\ \bibinfo {author} {\bibfnamefont
  {S.}~\bibnamefont {Koshimura}}\bibfield  {author} {  }\enquote {\bibinfo
  {title} {Enhancement of detecting permanent water and temporary water in
  flood disasters by fusing sentinel-1 and sentinel-2 imagery using deep
  learning algorithms: Demonstration of sen1floods11 benchmark datasets},}\ ,\
  \href@noop {} {\bibfield  {journal} {\bibinfo  {journal} {Remote Sensing}\
  }\textbf {\bibinfo {volume} {13}},\ p.\ \bibinfo {pages} {2220} (\bibinfo
  {year} {2021})}\BibitemShut {NoStop}%
\bibitem [{\citenamefont {Zhou}\ \emph {et~al.}(2018)\citenamefont {Zhou},
  \citenamefont {Rahman~Siddiquee}, \citenamefont {Tajbakhsh},\ and\
  \citenamefont {Liang}}]{zhou18unet++}%
  \BibitemOpen
  \bibfield  {author} {\bibinfo {author} {\bibfnamefont {Z.}~\bibnamefont
  {Zhou}}, \bibinfo {author} {\bibfnamefont {M.~M.}\ \bibnamefont
  {Rahman~Siddiquee}}, \bibinfo {author} {\bibfnamefont {N.}~\bibnamefont
  {Tajbakhsh}}, \ and\ \bibinfo {author} {\bibfnamefont {J.}~\bibnamefont
  {Liang}},\ }\enquote {\bibinfo {title} {{Unet++: A nested u-net architecture
  for medical image segmentation}},}\ in\ \href@noop {} {\emph {\bibinfo
  {booktitle} {Deep Learning in Medical Image Analysis and Multimodal Learning
  for Clinical Decision Support: 4th International Workshop, and 8th
  International Workshop, Held in Conjunction with MICCAI 2018, Proceedings
  4}}}\ (\bibinfo {organization} {Springer},\ \bibinfo {year} {2018})\ \unskip,
  pp.\ \bibinfo {pages} {3--11}\BibitemShut {NoStop}%
\bibitem [{\citenamefont {Chen}\ \emph {et~al.}(2018)\citenamefont {Chen},
  \citenamefont {Zhu}, \citenamefont {Papandreou}, \citenamefont {Schroff},\
  and\ \citenamefont {Adam}}]{chen18deeplabv3+}%
  \BibitemOpen
  \bibfield  {author} {\bibinfo {author} {\bibfnamefont {L.-C.}\ \bibnamefont
  {Chen}}, \bibinfo {author} {\bibfnamefont {Y.}~\bibnamefont {Zhu}}, \bibinfo
  {author} {\bibfnamefont {G.}~\bibnamefont {Papandreou}}, \bibinfo {author}
  {\bibfnamefont {F.}~\bibnamefont {Schroff}}, \ and\ \bibinfo {author}
  {\bibfnamefont {H.}~\bibnamefont {Adam}},\ }\enquote {\bibinfo {title}
  {{Encoder-decoder with atrous separable convolution for semantic image
  segmentation}},}\ in\ \href@noop {} {\emph {\bibinfo {booktitle} {Proceedings
  of the European conference on computer vision (ECCV)}}}\ (\bibinfo {year}
  {2018})\ \unskip, pp.\ \bibinfo {pages} {801--818}\BibitemShut {NoStop}%
\bibitem [{\citenamefont {Buslaev}\ \emph {et~al.}(2020)\citenamefont
  {Buslaev}, \citenamefont {Iglovikov}, \citenamefont {Khvedchenya},
  \citenamefont {Parinov}, \citenamefont {Druzhi-nin},\ and\ \citenamefont
  {Kalinin}}]{buslaev20albumentations}%
  \BibitemOpen
\bibinfo {author} {\bibfnamefont {A.}~\bibnamefont {Buslaev}}, \bibinfo
  {author} {\bibfnamefont {V.~I.}\ \bibnamefont {Iglovikov}}, \bibinfo {author}
  {\bibfnamefont {E.}~\bibnamefont {Khvedchenya}}, \bibinfo {author}
  {\bibfnamefont {A.}~\bibnamefont {Parinov}}, \bibinfo {author} {\bibfnamefont
  {M.}~\bibnamefont {Druzhi-nin}}, \ and\ \bibinfo {author} {\bibfnamefont
  {A.~A.}\ \bibnamefont {Kalinin}}\bibfield  {author} {  }\enquote {\bibinfo
  {title} {{Albumentations: fast and flexible image augmentations}},}\ ,\
  \href@noop {} {\bibfield  {journal} {\bibinfo  {journal} {Information}\
  }\textbf {\bibinfo {volume} {11}},\ p.\ \bibinfo {pages} {125} (\bibinfo
  {year} {2020})}\BibitemShut {NoStop}%
\bibitem [{\citenamefont {Alomar}, \citenamefont {Aysel},\ and\ \citenamefont
  {Cai}(2023)}]{alomar2023data}%
  \BibitemOpen
\bibinfo {author} {\bibfnamefont {K.}~\bibnamefont {Alomar}}, \bibinfo {author}
  {\bibfnamefont {H.~I.}\ \bibnamefont {Aysel}}, \ and\ \bibinfo {author}
  {\bibfnamefont {X.}~\bibnamefont {Cai}}\bibfield  {author} {  }\enquote
  {\bibinfo {title} {Data augmentation in classification and segmentation: A
  survey and new strategies},}\ ,\ \href@noop {} {\bibfield  {journal}
  {\bibinfo  {journal} {Journal of Imaging}\ }\textbf {\bibinfo {volume} {9}},\
  p.~\bibinfo {pages} {46} (\bibinfo {year} {2023})}\BibitemShut {NoStop}%
\bibitem [{\citenamefont {Deng}\ \emph {et~al.}(2009)\citenamefont {Deng},
  \citenamefont {Dong}, \citenamefont {Socher}, \citenamefont {Li},
  \citenamefont {Li},\ and\ \citenamefont {Fei-Fei}}]{deng09imagenet}%
  \BibitemOpen
  \bibfield  {author} {\bibinfo {author} {\bibfnamefont {J.}~\bibnamefont
  {Deng}}, \bibinfo {author} {\bibfnamefont {W.}~\bibnamefont {Dong}}, \bibinfo
  {author} {\bibfnamefont {R.}~\bibnamefont {Socher}}, \bibinfo {author}
  {\bibfnamefont {L.-J.}\ \bibnamefont {Li}}, \bibinfo {author} {\bibfnamefont
  {K.}~\bibnamefont {Li}}, \ and\ \bibinfo {author} {\bibfnamefont
  {L.}~\bibnamefont {Fei-Fei}},\ }\enquote {\bibinfo {title} {Imagenet: A
  large-scale hierarchical image database},}\ in\ \href@noop {} {\emph
  {\bibinfo {booktitle} {2009 IEEE conference on computer vision and pattern
  recognition}}}\ (\bibinfo {organization} {Ieee},\ \bibinfo {year} {2009})\
  \unskip, pp.\ \bibinfo {pages} {248--255}\BibitemShut {NoStop}%
\end{thebibliography}%

\end{document}